\documentclass{article}
\usepackage[utf8]{inputenc}
\usepackage{amsmath, amssymb, amsfonts, amsthm}
\usepackage{algorithm}
\usepackage{algpseudocode}
\usepackage{bm}
\usepackage{booktabs}
\usepackage{graphicx}
\usepackage{geometry}
\usepackage{microtype}
\usepackage{hyperref}
\usepackage{orcidlink}

\usepackage{xcolor} 
\definecolor{darkblue}{rgb}{0.0, 0.0, 0.55}

\usepackage[round, sort, authoryear]{natbib}

\newcommand{\R}{\mathbb{R}}

\newcommand{\Norm}{\mathcal{N}}
\newcommand{\bW}{\mathbf{W}}
\newcommand{\bx}{\mathbf{x}}
\newcommand{\by}{\mathbf{y}}
\newcommand{\bz}{\mathbf{z}}
\newcommand{\bmu}{\boldsymbol{\mu}}
\newcommand{\bsigma}{\boldsymbol{\sigma}}
\newcommand{\bepsilon}{\boldsymbol{\epsilon}}
\usepackage{graphicx}

\usepackage{tikz}
\usetikzlibrary{shapes.geometric, arrows.meta, positioning, calc, fit, backgrounds, shadows}
\usepackage{amsmath, amssymb}

\definecolor{myblue}{RGB}{70, 130, 180}
\definecolor{myred}{RGB}{220, 60, 60}
\definecolor{mygreen}{RGB}{60, 179, 113}
\definecolor{myorange}{RGB}{255, 165, 0}
\definecolor{mygray}{RGB}{240, 240, 240}
\definecolor{textgray}{RGB}{80, 80, 80}
\title{Hybrid Dual-Path Linear Transformations for Efficient Transformer Architectures}

\author{
  \textbf{Vladimer Khasia \orcidlink{0009-0002-3320-8142}} \\
  Independent Researcher \\
  \texttt{vladimer.khasia.1@gmail.com}
}
\date{February 05, 2026}

\begin{document}

\maketitle

\begin{abstract}
Standard Transformer architectures rely heavily on dense linear transformations, treating feature projection as a monolithic, full-rank operation. We argue that this formulation is inefficient and lacks the structural inductive bias necessary for distinguishing between local feature preservation and global context integration. To address this, we introduce the \textbf{Hybrid Dual-Path Linear (HDPL)} operator, which decomposes the affine transformation into two topologically distinct pathways: a sparse block-diagonal component for high-rank local processing, and a low-rank Variational Autoencoder (VAE) bottleneck for global context regularization. 

By ``surgically'' replacing specific projections (Query, Key, Value, Gate, Up) with HDPL operators while retaining standard dense layers for aggregation (Output, Down), we achieve a superior balance of efficiency and representational power. Experiments on the FineWeb-Edu dataset demonstrate that the HDPL architecture outperforms a standard Llama-style baseline, reducing validation loss while simultaneously reducing parameter count by $\approx 6.8\%$. Beyond immediate performance gains, we discuss how the explicit materialization of a probabilistic latent space within the Transformer backbone serves as a vital architectural affordance, offering new pathways for inference-time or hypernetwork induced control, continual adaptation, interpretability, and cross-model or cross-modal synchronization. 

The code is available at {\url{https://github.com/VladimerKhasia/HDPL}}

%%%%%%%%%%%%%%%%%%%%%%%%%%%%%%
%%%%%%%%%%%%%%%%%%%%%%%%%%%%%%
\begin{figure}[htbp]
    \centering
    \resizebox{\textwidth}{!}{% 
    
    \begin{tikzpicture}[
        node distance=1.5cm,
        font=\sffamily\small,
        >={Latex[width=2mm,length=2mm]},
        % Styles
        base/.style={draw, rounded corners=2pt, align=center, minimum height=0.8cm, drop shadow={opacity=0.15}},
        input/.style={base, fill=mygray, width=1.5cm},
        block/.style={base, fill=white, draw=textgray, line width=0.5pt},
        % STANDARD LINEAR (Blue) - Used for Wo and Wdown
        linear/.style={base, fill=myblue!10, draw=myblue, text=black, line width=0.8pt},
        % HYBRID LINEAR (Red) - Used for Wq, Wk, Wv, Wgate, Wup
        hybrid/.style={base, fill=myred!10, draw=myred, line width=1.2pt, text=black},
        op/.style={circle, draw=textgray, fill=white, inner sep=0pt, minimum size=0.6cm, drop shadow},
        loss/.style={base, dashed, fill=mygreen!10, draw=mygreen, text=mygreen!50!black, font=\footnotesize},
        connector/.style={->, draw=textgray, line width=0.8pt},
        loss_conn/.style={->, draw=mygreen, dashed, line width=0.6pt}
    ]
    
    % ==========================================================
    % PANEL A: The HybridDualPathLinear Micro-Architecture
    % ==========================================================
    
    % Input
    \node (x) [base, fill=gray!10] {Input $X$\\$[B, L, D]$};
    
    % --- Path 1: Block Diagonal (Detail) ---
    % CHANGED LABEL TO MATCH PYTHON CODE: .transpose(1, 2)
    \node (perm1) [block, above left=1.0cm and 0.4cm of x, anchor=south, font=\footnotesize] {Transpose (1, 2)\\$\to [B, D, L]$};
    
    \node (conv) [linear, above=0.5cm of perm1, font=\footnotesize] {Block-Diag Linear\\(Conv1d, Groups=$N$)};
    
    % CHANGED LABEL TO MATCH PYTHON CODE: .transpose(1, 2)
    \node (perm2) [block, above=0.5cm of conv, font=\footnotesize] {Transpose (1, 2)\\$\to [B, L, D]$};
    
    % --- Path 2: VAE (Context) ---
    \node (mu) [linear, above right=0.8cm and 0.1cm of x, minimum width=1.0cm, font=\footnotesize] {$\mu$\\Enc};
    \node (logvar) [linear, right=0.1cm of mu, minimum width=1.0cm, font=\footnotesize] {$\log\sigma^2$\\Enc};
    \node (sample) [op, above=0.6cm of mu, xshift=0.6cm] {$\mathcal{N}$};
    \node (act) [block, above=0.4cm of sample, minimum height=0.5cm, font=\footnotesize] {SiLU};
    \node (decoder) [linear, above=0.4cm of act, font=\footnotesize] {Decoder\\(Low Rank)};
    
    % JSD Loss (Internal to layer)
    \node (loss_internal) [loss, right=0.5cm of sample, align=center] {$\mathcal{L}_{JSD}$\\(Layer Loss)};
    
    % --- Merge ---
    \coordinate (merge_y) at ($(perm2.north)!0.5!(decoder.north) + (0, 0.6)$);
    \node (add) [op, at={(merge_y -| x)}] {+};
    \node (out) [base, fill=gray!10, above=0.5cm of add] {Output $Y$};
    
    % Connections Path 1
    \draw[connector] (x.north) -- ++(0,0.2) -| (perm1.south);
    \draw[connector] (perm1) -- (conv);
    \draw[connector] (conv) -- (perm2);
    \draw[connector] (perm2.north) |- (add.west) node[midway, above, font=\tiny, xshift=-0.2cm] {Detail Path};
    
    % Connections Path 2
    \draw[connector] (x.north) -- ++(0,0.2) -| (mu.south);
    \draw[connector] (x.north) -- ++(0,0.2) -| (logvar.south);
    \draw[connector] (mu) -- (sample);
    \draw[connector] (logvar) -- (sample);
    \draw[connector] (sample) -- (act) node[midway, right, font=\tiny] {$Z$};
    \draw[connector] (act) -- (decoder);
    \draw[connector] (decoder.north) |- (add.east) node[midway, above, font=\tiny] {Context Path};
    
    % Loss Connection
    \draw[loss_conn] (mu.east) -| (loss_internal.south);
    \draw[loss_conn] (logvar.east) -| (loss_internal.south);
    
    % Output Connection
    \draw[connector] (add) -- (out);
    
    % Background Box for Panel A
    \begin{scope}[on background layer]
        \node (panelA) [draw=gray!30, fill=white, dashed, rounded corners, fit=(x) (perm1) (loss_internal) (out), inner sep=8pt, label={[anchor=south east, font=\bfseries\color{gray}]south east:HybridDualPath}] {};
    \end{scope}
    
    % ==========================================================
    % PANEL B: Integration in Transformer
    % ==========================================================
    
    \coordinate (offset) at (8.5, -1.5); 
    
    % Input
    \node (emb) [base, fill=gray!10] at (offset) {Token Emb + Pos};
    \node (ln1) [block, above=0.5cm of emb] {RMSNorm};
    
    % -- Attention Block --
    \node (q) [hybrid, above left=0.6cm and -0.8cm of ln1, minimum width=0.8cm, font=\footnotesize] {$W_Q$};
    \node (k) [hybrid, right=0.2cm of q, minimum width=0.8cm, font=\footnotesize] {$W_K$};
    \node (v) [hybrid, right=0.2cm of k, minimum width=0.8cm, font=\footnotesize] {$W_V$};
    
    \node (attn) [block, above=0.6cm of k, minimum width=2.8cm, align=center, font=\footnotesize] {Rotary Embeddings\\+\\Scaled Dot-Product Attn};
    
    \node (proj_o) [linear, above=0.6cm of attn, minimum width=2.5cm] {$W_O$ (Standard)}; 
    
    \node (add1) [op, above=0.4cm of proj_o] {+};
    
    % -- MLP Block --
    \node (ln2) [block, above=0.5cm of add1] {RMSNorm};
    
    \node (gate) [hybrid, above left=0.6cm and -0.5cm of ln2, minimum width=1.2cm, font=\footnotesize] {$W_{Gate}$};
    \node (up) [hybrid, above right=0.6cm and -0.5cm of ln2, minimum width=1.2cm, font=\footnotesize] {$W_{Up}$};
    
    \node (mult) [op, above=0.8cm of ln2, font=\footnotesize] {$\times$};
    \node (silu_mlp) [block, above=0.2cm of gate, minimum height=0.4cm, font=\tiny] {SiLU};
    
    \node (proj_down) [linear, above=0.6cm of mult, minimum width=2.5cm] {$W_{Down}$ (Standard)}; 
    
    \node (add2) [op, above=0.4cm of proj_down] {+};
    \node (final) [base, fill=gray!10, above=0.5cm of add2] {To Next Layer};
    
    % -- Global Loss Aggregation Node --
    \node (total_loss) [loss, right=1.5cm of mult, align=center] {$\sum \mathcal{L}_{JSD}$\\(Total Aux Loss)};
    
    % Connections Panel B
    \draw[connector] (emb) -- (ln1);
    
    % Attn Internal
    \draw[connector] (ln1.north) -- ++(0,0.2) -| (q.south);
    \draw[connector] (ln1.north) -- ++(0,0.2) -- (k.south);
    \draw[connector] (ln1.north) -- ++(0,0.2) -| (v.south);
    \draw[connector] (q) -- (attn.south -| q);
    \draw[connector] (k) -- (attn.south -| k);
    \draw[connector] (v) -- (attn.south -| v);
    \draw[connector] (attn) -- (proj_o);
    \draw[connector] (proj_o) -- (add1);
    % Residual 1
    \draw[connector] (emb.west) -- ++(-0.6,0) |- (add1.west);
    
    % MLP Internal
    \draw[connector] (add1) -- (ln2);
    \draw[connector] (ln2.north) -- ++(0,0.2) -| (gate.south);
    \draw[connector] (ln2.north) -- ++(0,0.2) -| (up.south);
    \draw[connector] (gate) -- (silu_mlp);
    \draw[connector] (silu_mlp.north) -- (mult.west);
    \draw[connector] (up.north) |- (mult.east);
    \draw[connector] (mult) -- (proj_down);
    \draw[connector] (proj_down) -- (add2);
    % Residual 2
    \draw[connector] (add1.east) -- ++(1.2,0) |- (add2.east);
    
    \draw[connector] (add2) -- (final);
    
    % Loss Connections
    \draw[loss_conn] (v.east) -- ++(0.2,0) -| (total_loss.south) node[pos=0.2, below, font=\tiny] {from Q,K,V};
    \draw[loss_conn] (up.east) -- (total_loss.west);
    
    % Panel B Box
    \begin{scope}[on background layer]
        \node (panelB) [draw=gray!30, fill=white, dashed, rounded corners, fit=(emb) (final) (add1) (ln2) (total_loss), inner sep=15pt, label={[anchor=south east, font=\bfseries\color{gray}]south east:Surgical Integration}] {};
    \end{scope}
    
    % ==========================================================
    % LEGEND
    % ==========================================================
    
    \node [right=0.2cm of panelB.north east, anchor=north west, align=left, font=\footnotesize] (legend) {
        \textbf{Legend:}\\
        \tikz\node[linear, minimum width=0.3cm, minimum height=0.3cm, inner sep=0]{}; Standard Projection\\
        \tikz\node[hybrid, minimum width=0.3cm, minimum height=0.3cm, inner sep=0]{}; \textbf{HybridDualPath}\\
        \tikz\node[loss, minimum width=0.3cm, minimum height=0.3cm, inner sep=0]{}; Auxiliary Loss Flow
    };
    
    \node [above=0.1cm of panelA.north, font=\bfseries] {(a) Topology-Corrected Projection (Hybrid Layer)};
    \node [above=0.1cm of panelB.north, font=\bfseries] {(b) Transformer Integration};
    
    \end{tikzpicture}
    } % End resizebox
    \caption{Architecture Overview: (a) The proposed \textbf{HybridDualPath} layer decomposes projections into a sparse detail path and a global VAE context path. (b) The integration within the Transformer block, where $W_Q, W_K, W_V$ (Attention) and $W_{Gate}, W_{Up}$ (MLP) are replaced by our hybrid layer (Red), while $W_O$ and $W_{Down}$ remain standard (Blue) to aggregate features.}
\end{figure}
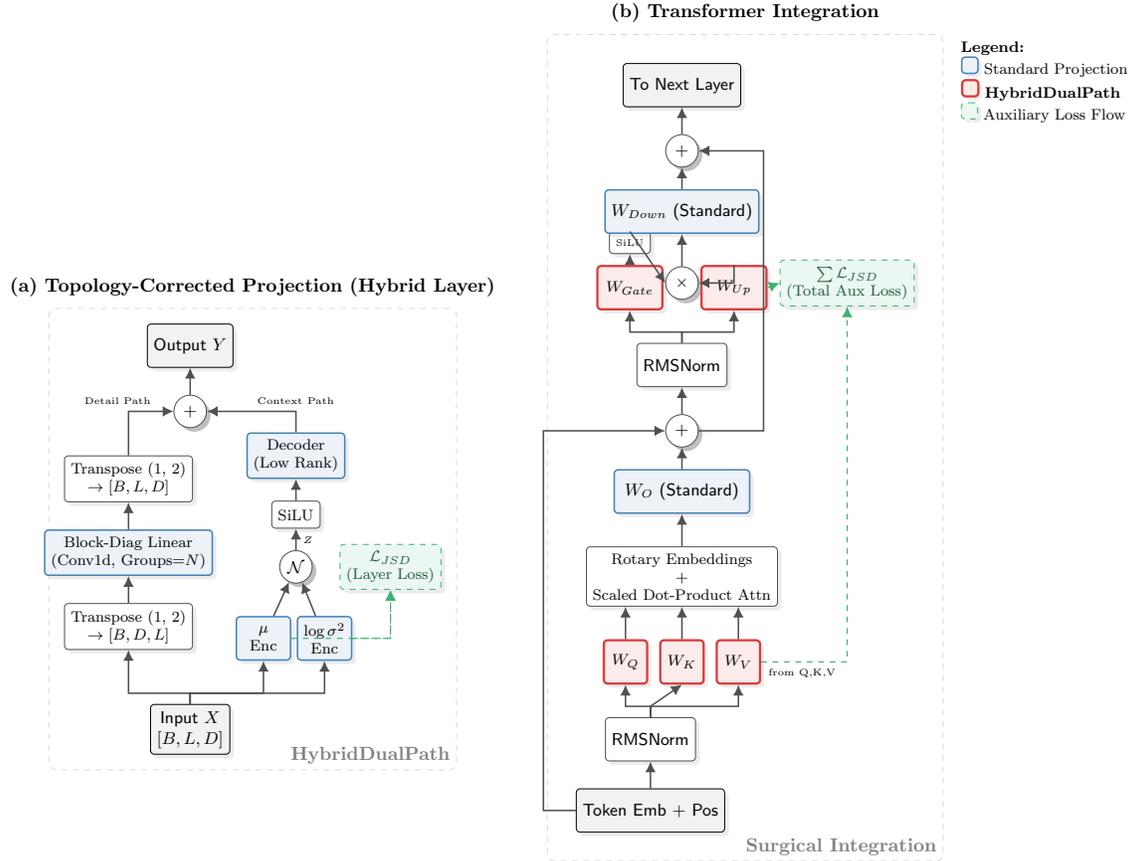
%%%%%%%%%%%%%%%%%%%%%%%%%%%%%%
%%%%%%%%%%%%%%%%%%%%%%%%%%%%%%

\end{abstract}
\section{Introduction}

The Transformer architecture \citep{NIPS2017_3f5ee243} has established itself as the de facto standard for sequence modeling tasks, driven largely by the predictable scalability of its performance with respect to parameter count and compute \citep{kaplan2020scalinglawsneurallanguage, hoffmann2022trainingcomputeoptimallargelanguage}. However, this scaling relies heavily on the stacking of dense linear transformations—specifically within the Multi-Head Attention (MHA) and Feed-Forward Network (FFN) blocks—which constitute the vast majority of the model's parameters and computational footprint (FLOPs).

Standard approaches to mitigating this computational burden often involve post-hoc compression techniques such as quantization \citep{dettmers2022llmint88bitmatrixmultiplication}, pruning \citep{JMLR:v22:21-0366}, or low-rank adaptation \citep{hu2021loralowrankadaptationlarge}. While effective, these methods typically view the weight matrix $\bW \in \R^{d_{out} \times d_{in}}$ as a monolithic entity to be compressed uniformly. We argue that this view ignores the distinct topological roles played by linear projections: preserving local feature identity (high-rank, sparse) versus integrating global context (low-rank, dense). Imposing a low-rank bottleneck uniformly across a layer risks collapsing the high-frequency feature details necessary for precise token representation, while unstructured sparsity fails to capture the semantic correlations inherent in language.

In this work, we propose a \textit{Surgical Hybrid Architecture} that fundamentally rethinks the parametrization of the linear layer. We introduce the \textbf{Hybrid Dual-Path Linear (HDPL)} operator, which decomposes the affine transformation into two topologically distinct pathways:
\begin{enumerate}
    \item A \textbf{Local Detail Path} parametrized by block-diagonal sparsity, ensuring high-rank capacity for maintaining feature specificity without cross-group interference.
    \item A \textbf{Global Context Path} parametrized by a low-rank Variational Autoencoder (VAE) \citep{kingma2022autoencodingvariationalbayes}, which acts as a stochastic information bottleneck to capture and regularize global dependencies.
\end{enumerate}

Unlike previous structured matrix approaches that rely on fixed patterns (e.g., Monarch matrices \citep{pmlr-v162-dao22a}), our approach integrates a probabilistic regularizer—the Bounded Kullback-Leibler (KL) divergence—directly into the forward pass of the linear layer. This enforces a disentangled latent representation of the global context, effectively filtering noise during the training process.

We evaluate this methodology by surgically replacing the Query ($W_q$), Key ($W_k$), Value ($W_v$), Gate ($W_{gate}$), and Up ($W_{up}$) projections in a Llama-style Transformer \citep{touvron2023llamaopenefficientfoundation}, while retaining standard dense projections for the Output ($W_o$) and Down ($W_{down}$) layers to preserve aggregation capacity. Training on the FineWeb-Edu dataset \citep{penedo2024the}, our experiments demonstrate that the HDPL-enhanced model outperforms a standard dense baseline while reducing the parameter count by approximately $6.8\%$.

Our contributions are as follows:
\begin{itemize}
    \item We formalize the \textbf{Hybrid Dual-Path Linear (HDPL)} operator, a mathematically rigorous replacement for dense layers that combines block-diagonal sparsity with variational low-rank approximation.
    \item We provide a \textbf{Surgical Integration Strategy} that identifies which Transformer sub-components benefit from hybrid topology versus those that require full-rank density.
    \item We provide empirical evidence that integrating VAE bottlenecks into the Transformer backbone acts as a superior inductive bias for language modeling, yielding better convergence properties than fully deterministic over-parameterized baselines.
\end{itemize}

\section{Methodology}
\label{sec:method}
We propose a surgically applied modification to the linear projections within the Transformer architecture. We replace standard dense linear layers with a \textbf{Hybrid Dual-Path Linear (HDPL)} operator. This operator decomposes the transformation of the input representations into two distinct topological paths: a \textit{High-Rank Local Path} utilizing block-diagonal sparsity to preserve feature specificity, and a \textit{Low-Rank Variational Path} utilizing a stochastic bottleneck to capture global context.

\subsection{The Hybrid Dual-Path Operator}

Let $\bx \in \R^{B \times L \times D_{in}}$ denote an input tensor, where $B$ is the batch size, $L$ is the sequence length, and $D_{in}$ is the input dimension. A standard linear layer computes $\by = \bx \bW^\top$, where $\bW \in \R^{D_{out} \times D_{in}}$. The HDPL operator, denoted as $\mathcal{H}(\bx)$, approximates this transformation via the sum of a sparse local term and a dense global term:
\begin{equation}
    \mathcal{H}(\bx) = \underbrace{\Psi_{\text{local}}(\bx)}_{\text{Detail Path}} + \underbrace{\Phi_{\text{global}}(\bx)}_{\text{Context Path}}
\end{equation}

\subsubsection{Path 1: Block-Diagonal Local Projection}
The local path, $\Psi_{\text{local}}$, is designed to maintain high-rank transformations within local feature subspaces while reducing parameter count. We partition the input dimension $D_{in}$ and output dimension $D_{out}$ into $K$ disjoint blocks (groups). Assuming $D_{in}$ and $D_{out}$ are divisible by $K$, the weight matrix $\bW^{(B)}$ takes a block-diagonal form:
\begin{equation}
    \bW^{(B)} = \text{diag}(\bW_1, \bW_2, \dots, \bW_K), \quad \text{where } \bW_k \in \R^{\frac{D_{out}}{K} \times \frac{D_{in}}{K}}
\end{equation}
The operation is efficiently implemented via grouped 1D convolution. For an input vector $\bx_t$ at time step $t$, the transformation is:
\begin{equation}
    \Psi_{\text{local}}(\bx_t) = \bx_t (\bW^{(B)})^\top
\end{equation}
This path preserves the "identity" and high-frequency details of the signal by preventing cross-group interference, effectively modeling independent feature subspaces.

\subsubsection{Path 2: Variational Global Context}
To compensate for the lack of cross-group communication in the local path, we introduce a global context path $\Phi_{\text{global}}$ parameterized as a Variational Autoencoder (VAE) bottleneck. This path projects the input into a low-dimensional latent space $\mathcal{Z} \subset \R^{R}$, where $R \ll D_{in}$ is the rank.

The encoding stage predicts the parameters of the posterior distribution $q(\bz|\bx) = \Norm(\bmu, \text{diag}(\bsigma^2))$:
\begin{align}
    \bmu &= \bx \bW_{\mu}^\top, \quad \bW_{\mu} \in \R^{R \times D_{in}} \\
    \log \bsigma^2 &= \bx \bW_{\sigma}^\top, \quad \bW_{\sigma} \in \R^{R \times D_{in}}
\end{align}
During training, we utilize the reparameterization trick to sample $\bz$:
\begin{equation}
    \bz = \bmu + \bsigma \odot \bepsilon, \quad \bepsilon \sim \Norm(\mathbf{0}, \mathbf{I})
\end{equation}
During inference, the operation is deterministic, setting $\bz = \bmu$. The latent representation $\bz$ is then passed through a non-linearity $\sigma(\cdot)$ (specifically SiLU) and projected back to the output dimension via a decoder $\bW_{dec} \in \R^{D_{out} \times R}$:
\begin{equation}
    \Phi_{\text{global}}(\bx) = \bW_{dec} \sigma(\bz)
\end{equation}

\subsection{Algorithm Specification}

The forward pass of the HDPL layer, including the auxiliary loss calculation, is formally specified in Algorithm \ref{alg:hdpl}.

\begin{algorithm}[h]
\caption{Hybrid Dual-Path Linear (HDPL) Forward Pass}
\label{alg:hdpl}
\begin{algorithmic}[1]
\Require Input $\bx \in \R^{B \times L \times D_{in}}$, Rank $R$, Groups $K$, Scaling $\beta$
\Ensure Output $\by$, Auxiliary Loss $\mathcal{L}_{aux}$

\State \textbf{Path 1: Local (Block-Diagonal)}
\State $\by_{\text{local}} \leftarrow \text{GroupedConv1D}(\bx, \text{groups}=K)$ \Comment{Eq. 3}

\State \textbf{Path 2: Global (Variational)}
\State $\bmu \leftarrow \text{Linear}_{\mu}(\bx)$
\State $\log\mathbf{v} \leftarrow \text{Linear}_{\sigma}(\bx)$
\If{Training}
    \State $\bsigma \leftarrow \exp(0.5 \cdot \log\mathbf{v})$
    \State $\bepsilon \sim \Norm(\mathbf{0}, \mathbf{I})$
    \State $\bz \leftarrow \bmu + \bsigma \odot \bepsilon$
    \State $\mathcal{L}_{aux} \leftarrow \text{BoundedKL}(\bmu, \log\mathbf{v}) \times \beta$
\Else
    \State $\bz \leftarrow \bmu$
    \State $\mathcal{L}_{aux} \leftarrow 0$
\EndIf
\State $\by_{\text{global}} \leftarrow \text{Linear}_{dec}(\text{SiLU}(\bz))$

\State \textbf{Combine}
\State $\by \leftarrow \by_{\text{local}} + \by_{\text{global}}$
\State \Return $\by, \mathcal{L}_{aux}$
\end{algorithmic}
\end{algorithm}

\subsection{Training Objective and Regularization}

The training objective is augmented with a regularization term derived from the Kullback-Leibler (KL) divergence between the approximate posterior $q(\bz|\bx)$ and a standard normal prior $p(\bz) = \Norm(\mathbf{0}, \mathbf{I})$. To ensure training stability and prevent posterior collapse, we employ a Bounded KL objective acting as a proxy for the Geometric Jensen-Shannon Divergence:

\begin{equation}
    \mathcal{L}_{KL} = -\frac{1}{2} \sum_{j=1}^{R} \left( 1 + \log(\bsigma_j^2) - \bmu_j^2 - \bsigma_j^2 \right)
\end{equation}
\begin{equation}
    \mathcal{L}_{aux} = \beta \cdot \frac{1}{N} \sum_{i=1}^{N} \text{clamp}(\mathcal{L}_{KL}^{(i)}, \text{max}=\ln 2)
\end{equation}
where $\beta$ is a hyperparameter balancing reconstruction fidelity and latent space regularization. The total loss for the network is $\mathcal{L}_{total} = \mathcal{L}_{CE} + \sum_{l \in \mathcal{S}} \mathcal{L}_{aux}^{(l)}$, where $\mathcal{S}$ is the set of hybrid layers.

\subsection{Surgical Integration into Transformer}

We define a configuration set $\mathcal{C}_{hybrid} = \{W_q, W_k, W_v, W_{gate}, W_{up}\}$ representing the linear projections eligible for hybrid replacement. The output projection $W_o$ and the MLP down-projection $W_{down}$ retain standard dense formulations to ensure full-rank capacity at the aggregation points of the Attention and FFN blocks, respectively.

For a given attention head $h$, the query, key, and value projections become:
\begin{equation}
    Q_h, \mathcal{L}_Q = \mathcal{H}_q(\bx), \quad K_h, \mathcal{L}_K = \mathcal{H}_k(\bx), \quad V_h, \mathcal{L}_V = \mathcal{H}_v(\bx)
\end{equation}
Similarly, in the Feed-Forward Network (FFN) using SiLU gating:
\begin{align}
    \mathbf{g}, \mathcal{L}_g &= \mathcal{H}_{gate}(\bx) \\
    \mathbf{u}, \mathcal{L}_u &= \mathcal{H}_{up}(\bx) \\
    \bx_{ffn} &= W_{down}(\text{SiLU}(\mathbf{g}) \odot \mathbf{u})
\end{align}

\subsection{Complexity Analysis}
The parameter complexity of a standard linear layer is $\mathcal{O}(D_{in}D_{out})$. The HDPL operator reduces this to:
\begin{equation}
    \mathcal{O}\left(\frac{D_{in}D_{out}}{K} + 2 \cdot D_{in}R + D_{out}R\right)
\end{equation}
Given that $K \ge 1$ and $R \ll D_{in}$, the HDPL operator significantly reduces memory footprint and FLOPs while theoretically maintaining the capacity to model both high-frequency local dependencies (via the block diagonal term) and low-frequency global dependencies (via the low-rank term).

\section{Experiments}
\label{sec:experiments}

We evaluate the efficacy of the proposed Hybrid Dual-Path Linear (HDPL) operator  (Section \ref{sec:method}) by integrating it into a standard Transformer language model. We focus on a "Surgical" configuration where the HDPL operator replaces the Query ($W_q$), Key ($W_k$), Value ($W_v$), Gate ($W_{gate}$), and Up ($W_{up}$) projections. Consistent with the aggregation theory of Transformers, we retain the standard dense formulation for the Output ($W_o$) and Down ($W_{down}$) projections to preserve full-rank feature mixing at the summation nodes of the residual stream.

\subsection{Experimental Setup}

\textbf{Dataset.} We train all models on the \texttt{FineWeb-Edu} dataset \citep{penedo2024the}, a high-quality educational subset of the Common Crawl. The dataset is tokenized using the \texttt{SmolLM2} tokenizer \citep{allal2025smollm2smolgoesbig} with a vocabulary size of $V=49,152$. We utilize a streaming configuration with a sequence length of $L=2048$.

\textbf{Architecture.} We compare two architectures:
\begin{enumerate}
    \item \textbf{Baseline Transformer:} 
    
    A standard Llama-style architecture using RMSNorm, Rotary Positional Embeddings (RoPE), and SwiGLU activations. All linear layers are standard dense matrices.
    \item \textbf{Surgical Hybrid Transformer:} 
    
    Identical architecture, but with $\mathcal{C}_{hybrid} = \{W_q, W_k, W_v, W_{gate}, W_{up}\}$ replaced by HDPL operators. The latent rank is set to $R=128$ and the scaling factor $\beta=0.001$.
\end{enumerate}
Both models share the same depth ($N=4$ layers), model dimension ($D=512$), and attention heads ($H=8$).

\textbf{Training Protocol.} Models are trained for $20,000$ steps using the AdamW optimizer with $\beta_1=0.9, \beta_2=0.95$. We employ a cosine learning rate schedule with a warmup of $1000$ steps and a peak learning rate of $8 \times 10^{-4}$ (see Appendix \ref{app:hyperparams} for full details).

\subsection{Results and Analysis}

\textbf{Parameter Efficiency and Compression.}
As detailed in Table \ref{tab:model_comparison}, the Surgical Hybrid method reduces the total parameter count from $67.1$M to $62.5$M, representing a $\approx 6.8\%$ reduction in total model size (including embeddings). When considering only the transformed layers, the compression ratio is significantly higher. Despite this reduction in capacity, the Hybrid model demonstrates superior representational power.

\begin{table}[h!]
\centering
\caption{Comparison of Baseline vs. Surgical Hybrid Method. Loss values are reported on the validation split of FineWeb-Edu. The Hybrid method achieves lower perplexity with fewer parameters.}
\label{tab:model_comparison}
\begin{tabular}{lcccc}
\toprule
\textbf{Model} & \textbf{Params (M)} & \textbf{Size (MB)} & \textbf{Final Val Loss} & \textbf{Min Val Loss} \\
\midrule
Baseline (Dense) & 67.11 & 256.02 & 4.3206 & 4.3206 \\
\textbf{Surgical Hybrid (Ours)} & \textbf{62.53} & \textbf{238.54} & \textbf{4.2838} & \textbf{4.2838} \\
\bottomrule
\end{tabular}
\end{table}

\begin{figure}[h]
    \centering
    \includegraphics[width=0.9\linewidth]{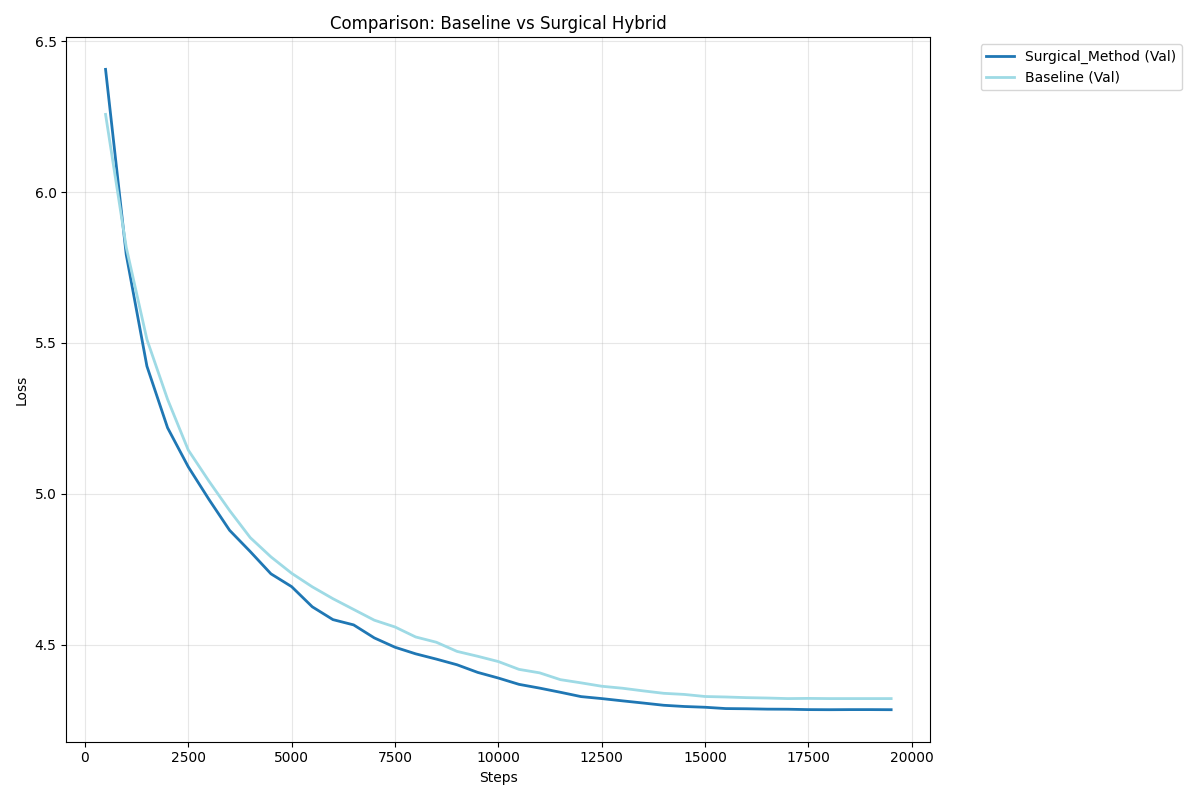}
    \caption{\textbf{Validation Loss Trajectories.} Comparison of the Baseline (Light Blue) and Surgical Hybrid (Dark Blue) models over 20,000 training steps. The Surgical Hybrid method exhibits strictly better sample efficiency, achieving lower loss values earlier in training and converging to a superior optimum (4.28 vs 4.32) despite having fewer parameters.}
    \label{fig:loss_curves}
\end{figure}

\textbf{Convergence Dynamics.}
Figure \ref{fig:loss_curves} illustrates the validation loss trajectories. The Baseline model follows a standard convergence curve, plateauing near a loss of $4.32$ around step 17,000. Conversely, the Surgical Hybrid model consistently outperforms the baseline after the initial warmup phase. By step 10,000, the Hybrid model reaches a loss of $4.38$, while the Baseline is at $4.44$. The Hybrid model continues to improve, reaching a final validation loss of $4.284$.

This result suggests that the dual-path topology—combining sparse high-rank local operations with a low-rank global variational bottleneck—acts as an effective inductive bias. The auxiliary loss $\mathcal{L}_{aux}$, which stabilizes the VAE path, remains negligible throughout training (it increases if out and down projections are replaced with hybrid architecture as well), indicating that the regularization term effectively shapes the latent space without dominating the gradient signal.

\textbf{Computational Throughput.}
We observe a trade-off in wall-clock throughput. The Baseline model processes $\approx 480$k tokens/sec, while the Hybrid model processes $\approx 270$k tokens/sec on TPU v3 hardware. This discrepancy is attributed to the current implementation of the HDPL operator, which relies on grouped convolutions and VAE sampling in PyTorch, lacking the highly fused kernel optimizations available for standard dense matrix multiplications (GEMM) on XLA devices. However, the increased information density (loss decrease per parameter) suggests that with kernel fusion, the HDPL operator could offer a favorable Pareto frontier for inference-constrained environments.

\section{Discussion}
\label{sec:discussion}

Our results demonstrate that the dense linear transformation, a ubiquitous primitive in Transformer architectures, acts as an over-parameterized upper bound on the necessary complexity for feature projection. By decomposing the transformation into a sparse, high-rank deterministic path and a low-rank stochastic path, the proposed Hybrid Dual-Path Linear (HDPL) operator achieves a superior validation loss ($4.28$ vs. $4.32$) with significantly fewer parameters ($62.5$M vs $67.1$M).

\subsection{Architectural Implications of the Dual-Path Topology}

The convergence properties observed in Figure \ref{fig:loss_curves} suggest that the inductive bias of the HDPL operator aligns better with the underlying structure of language data than standard affine transformations. We posit that the block-diagonal path ($\Psi_{\text{local}}$) effectively preserves local manifold geometry and high-frequency feature identity, preventing the "oversmoothing" often associated with low-rank approximations. Simultaneously, the Variational Autoencoder path ($\Phi_{\text{global}}$) captures global correlations.

The superior generalization of the HDPL model indicates that the imposition of a Kullback-Leibler divergence penalty ($\mathcal{L}_{aux}$) acts not merely as a regularizer, but as an information bottleneck. This forces the global path to learn a compressed, semantic representation of the input features, filtering out noise that a full-rank matrix might otherwise overfit.

\subsection{The Strategic Utility of the VAE Bottleneck}

Beyond the immediate gains in parameter efficiency, the incorporation of a variational bottleneck within the Transformer layers introduces distinct functional capabilities absent in standard architectures. We identify three key areas where this architectural choice serves as a foundational enabler for future research.

\subsubsection{Inference-Time Adaptability and Control}
The explicit materialization of a latent variable $\bz \sim q(\bz|\bx)$ provides a handle for inference-time intervention that is unavailable in deterministic networks.
\begin{enumerate}
    \item \textbf{Latent Manipulation:} Unlike activation patching, which requires modifying high-dimensional vectors, $\bz$ exists in a low-dimensional, regularized space ($R=128$). This enables fine-grained control over model output via latent arithmetic, clamping or conditioning. 
    \item \textbf{Continual Learning:} The VAE framework allows for the control of certainty, novelty and usefulness of the information as well as the  application of replay-based alignment methods on the latent distribution $p(\bz)$ rather than the weights. This offers a potential pathway to mitigate catastrophic forgetting by enforcing that the distribution of latents for new tasks remains consistent with the prior established by previous tasks.
\end{enumerate}

\subsubsection{Hypernetworks and Meta-Learning}
Controlling the weights of a standard Transformer via Hypernetworks is computationally intractable due to the sheer size of the target matrices (e.g., generating a $4096 \times 4096$ matrix requires predicting $16$M parameters).
The HDPL architecture drastically reduces the dimensionality of the target space. A Hypernetwork would only need to predict the parameters of the encoder/decoder interfaces ($\R^{D \times R}$) or just inject the generated latents, which are orders of magnitude smaller. This feasibility opens the door for dynamic, adaptable and high capacity networks.

\subsubsection{Cross-System Synchronization and alignment}
The probabilistic nature of the bottleneck ($\bz \sim \Norm(\bmu, \bsigma)$) provides a standardized protocol for alignment that deterministic weights lack.
\begin{itemize}
    \item \textbf{Federated Learning:} In distributed settings, averaging weights often leads to suboptimal performance due to permutation symmetries. Aligning the moments ($\bmu, \bsigma$) of the latent distributions across agents offers a robust alternative for knowledge distillation and aggregation.
    \item \textbf{Cross-Modal Alignment:} The bottleneck allows for the synchronization of disparate modalities (e.g., text and image embeddings) by forcing them to map to a shared latent topology defined by the prior $p(\bz)$. The HDPL operator thus acts as a native interface for multi-modal fusion without requiring massive adapter layers.
\end{itemize}

\subsection{limitations}
While the HDPL operator improves parameter efficiency and loss, our current implementation utilizing standard PyTorch primitives results in lower wall-clock throughput ($\approx 44\%$ reduction) compared to highly optimized GEMM kernels on TPU hardware. Future work must focus on writing custom fused kernels (e.g., in Triton or Pallas) to merge the block-diagonal and VAE operations, ensuring that the theoretical FLOPs reduction translates into realized speedups.

\section{Conclusion}

In this work, we have presented the Hybrid Dual-Path Linear (HDPL) operator, a method that challenges the assumption that dense, full-rank matrix multiplications are the irreducible atomic unit of Transformer computation. By surgically decomposing the linear projections of the Attention ($W_q, W_k, W_v$) and Feed-Forward ($W_{gate}, W_{up}$) blocks into a sparse local path and a variational global path, we demonstrated that it is possible to improve representational capacity while simultaneously reducing parameter count.

Our experiments on the FineWeb-Edu dataset reveal that the HDPL-enhanced architecture achieves a strictly superior validation loss compared to a standard dense baseline, despite a $\approx 6.8\%$ reduction in parameters. This result suggests that the dual-path topology provides a more effective inductive bias for language modeling: the block-diagonal component preserves the high-frequency identity of local features, while the VAE bottleneck forces the model to learn a disentangled, noise-robust representation of global context.

While our method is information-theoretically superior, its current implementation via standard PyTorch primitives incurs a latency penalty on modern hardware optimized for dense GEMM operations. The transition from theoretical advantage to practical wall-clock speedup will require the development of custom hardware-aware kernels that can fuse the sparse and stochastic operations.

Ultimately, the incorporation of a variational information bottleneck within the linear layer offers more than just compression; it introduces a probabilistic affordance for control, interpretation, and alignment. We hope this work serves as a foundational step toward a new class of methods that prioritize topological correctness over brute-force scale.

\bibliographystyle{plainnat} 
\bibliography{references}

\newpage
\appendix

\section{Appendix}

\subsection{Experimental Hyperparameters}
\label{app:hyperparams}

All experiments were conducted with the hyperparameters detailed in Table \ref{tab:hyperparams}. To ensure reproducibility, we utilized a fixed random seed of 42.

\begin{table}[h]
\centering
\caption{Hyperparameter configuration for the Surgical Hybrid and Baseline experiments.}
\label{tab:hyperparams}
\begin{tabular}{ll}
\toprule
\textbf{Hyperparameter} & \textbf{Value} \\
\midrule
\multicolumn{2}{c}{\textit{Architecture}} \\
Model Dimension ($D_{model}$) & 512 \\
Hidden Dimension (MLP) & 2048 \\
Number of Layers & 4 \\
Number of Heads & 8 \\
Head Dimension & 64 \\
Vocab Size & 49,152 \\
Sequence Length & 2048 \\
\midrule
\multicolumn{2}{c}{\textit{Optimization}} \\
Optimizer & AdamW \\
Learning Rate & $8 \times 10^{-4}$ \\
Min Learning Rate & $8 \times 10^{-5}$ \\
Weight Decay & 0.1 \\
Warmup Steps & 1000 \\
Max Steps & 20,000 \\
Batch Size (Global) & 32 \\
\midrule
\multicolumn{2}{c}{\textit{Hybrid Specifics}} \\
$\mathcal{C}_{hybrid}$ & $\{W_q, W_k, W_v, W_{gate}, W_{up}\}$ \\
Latent Rank ($R$) & 128 \\
VAE $\beta$ & 0.001 \\
Block Groups ($K$) & 8 \\
\bottomrule
\end{tabular}
\end{table}

\subsection{Hardware and Infrastructure}
Training was performed on a TPU VM provided by Kaggle utilizing the PyTorch XLA backend.

\subsection{Auxiliary Loss Behavior}
The auxiliary loss $\mathcal{L}_{aux}$ is a weighted geometric JSD proxy. In our experiments, we observed that $\mathcal{L}_{aux}$ did not collapse to zero, nor did it explode. It maintained a small magnitude relative to the Cross Entropy (CE) loss. 

\end{document}